\documentclass[letterpaper, 10 pt, conference]{ieeeconf}  % Comment this line out if you need a4paper

\IEEEoverridecommandlockouts                              % This command is only needed if 
\usepackage{graphics, graphicx} % for pdf, bitmapped graphics files
\usepackage{xcolor}
\usepackage{hyperref}
\usepackage{ulem}
\usepackage{amsmath} % assumes amsmath package installed
\usepackage{amsfonts} 
\usepackage{hhline}
\usepackage{dsfont}
\DeclareMathOperator*{\argmax}{arg\,max}

\title{\LARGE \bf
ConSOR: A Context-Aware Semantic Object Rearrangement Framework for Partially Arranged Scenes
}

\author{Kartik Ramachandruni, Max Zuo, and Sonia Chernova
\thanks{Georgia Institute of Technology, Atlanta, Georgia, United States.
Contact: \texttt{\{kvr6,zuo,chernova\}@gatech.edu}
}
}

\begin{document}
\raggedbottom

\maketitle
\thispagestyle{empty}
\pagestyle{empty}

\begin{figure*}[h]
    \centering
    \includegraphics[width=0.85\textwidth]{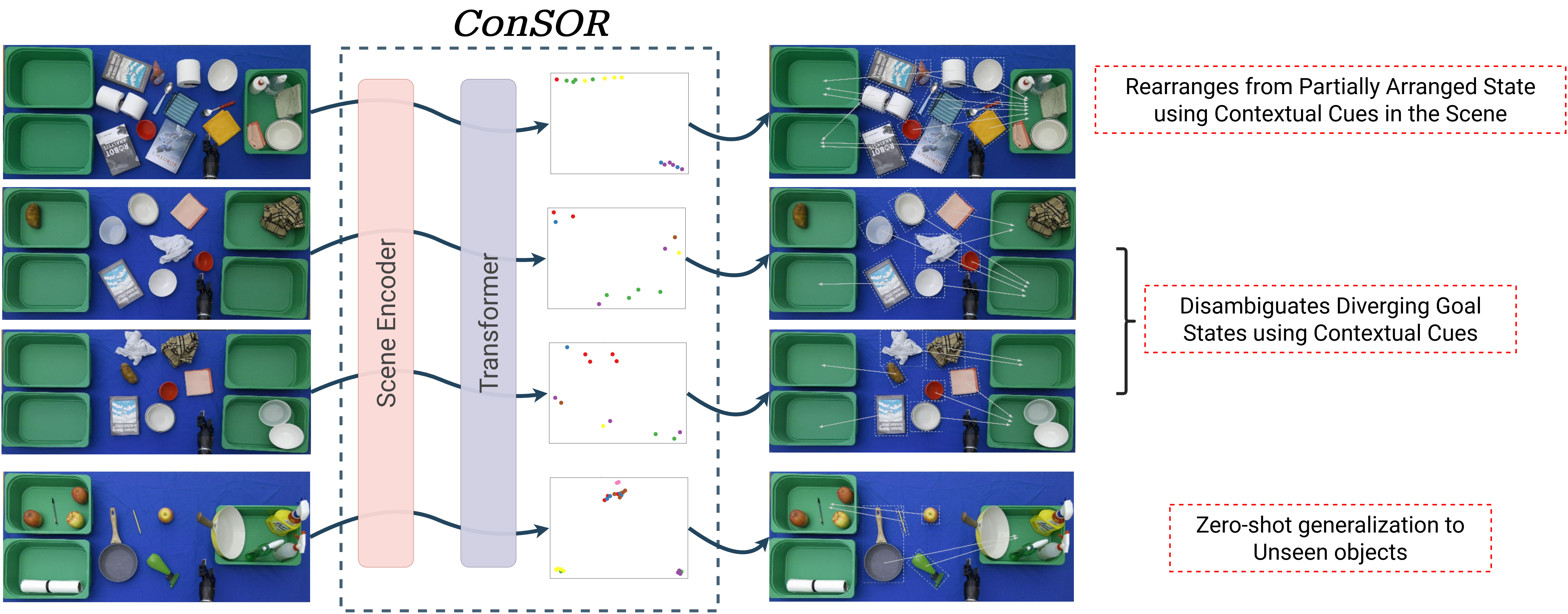}
    \caption{Our semantic object rearrangement framework, ConSOR, takes partially arranged object scenes (left), uses a Transformer-based neural architecture to infer contextual cues about the likely goal arrangement, and generates the desired arrangement state (right).  }
    \label{fig:summary}
\end{figure*}
%%%%%%%%%%%%%%%%%%%%%%%%%%%%%%%%%%%%%%%%%%%%%%%%%%%%%%%%%%%%%%%%%%%%%%%%%%%%%%%%
\begin{abstract}

Object rearrangement is the problem of enabling a robot to identify the correct object placement in a complex environment.  Prior work on object rearrangement has explored a diverse set of techniques for following user instructions to achieve some desired goal state.  Logical predicates, images of the goal scene, and natural language descriptions have all been used to instruct a robot in how to arrange objects. In this work, we argue that burdening the user with specifying goal scenes is not necessary in partially-arranged environments, such as common household settings.  Instead, we show that contextual cues from partially arranged scenes (i.e., the placement of some number of pre-arranged objects in the environment) provide sufficient context to enable robots to perform object rearrangement \textit{without any explicit user goal specification}. We introduce ConSOR, a Context-aware Semantic Object Rearrangement framework that utilizes contextual cues from a partially arranged initial state of the environment to complete the arrangement of new objects, without explicit goal specification from the user. We demonstrate that ConSOR strongly outperforms two baselines in generalizing to novel object arrangements and unseen object categories. The code and data are available at \href{https://github.com/kartikvrama/consor}{https://github.com/kartikvrama/consor}.

\end{abstract}

%%%%%%%%%%%%%%%%%%%%%%%%%%%%%%%%%%%%%%%%%%%%%%%%%%%%%%%%%%%%%%%%%%%%%%%%%%%%%%%%

\section{Introduction}

Consider a service robot tasked with putting away newly delivered groceries, or cleaning a living room. In both tasks, the environment is most likely already partially arranged, and that arrangement provides valuable clues for where new items should be placed.  For example, the pantry may already contain unfinished boxes of cereals and pasta on different shelves, while the left drawer of the refrigerator may contain half-finished vegetables.  Thus, new items, such as a box of oatmeal, should be placed in accordance with the user's existing organization scheme (e.g., near the cereal). Similarly, a book may naturally be placed alongside other books on the shelf rather than next to houseplants.

The general problem of identifying the correct item placement in a complex environment is known as the \textit{object rearrangement problem} \cite{batra2020rearrangement}.  Prior work on object rearrangement has explored a diverse set of techniques for following user instructions to achieve some desired goal state.  Logical predicates~\cite{paxton2022predicting,wang2022generalizable}, images of the goal scene~\cite{goyal2022ifor, wu2022transporters, goodwin2022semantically}, and natural language descriptions~\cite{liu2022structdiffusion, shridhar2022perceiver, shridhar2022cliport} have all been used to instruct a robot in how to arrange objects. However, all of the above techniques place a burden on the user to explicitly describe the goal state, or else to explicitly demonstrate the rearrangement task so that the robot can learn from demonstrations \cite{jain_transformers_2022, kapelyukh_my_2022}. 

In this work, we posit that contextual cues from partially arranged scenes (i.e., the placement of some number of pre-arranged objects in the environment) provide sufficient context to enable robots to perform object rearrangement \textit{without any explicit user goal specification}.  
Closely related to our work are those of Abdo et al.~\cite{abdo_organizing_2016} and Wu et al.~\cite{wu_targf_2022}, which reason about object similarities by learning object relationships from demonstrations of arranged environments, which are then generalized to novel environments. However, these works require that the desired organizational style in the goal state be known \textit{a priori} (e.g., specified by the user) instead of inferring this style from scene context. 

We introduce ConSOR, a Context-aware Semantic Object Rearrangement framework that utilizes contextual cues from a partially arranged initial state of the environment to complete the arrangement of new objects, without explicit goal specification from the user.  Figure~\ref{fig:summary} presents an overview of our framework. ConSOR reasons about the semantic properties of objects in the environment, and the context provided by the number of containers and existing placement of objects into containers, to infer the desired placement for new, unarranged objects. 
Additionally, ConSOR leverages prior commonsense knowledge from pre-trained ConceptNet embeddings to perform zero-shot generalization to scenes with objects unseen during training. Our work makes the following contributions:
\begin{itemize}
    \item We formalize the problem of object rearrangement in partially arranged environments.
    \item We present ConSOR, a Context-aware Semantic Object Rearrangement framework that replaces human instruction with contextual cues from the initial state of the environment to infer the desired goal state of an object rearrangement task.
    \item We contribute a dataset of~8k rearranged goal states from a dataset of $38$ household objects, with each goal state associated with one of four predefined organizational \textit{schemas}.
    \item We demonstrate that ConSOR is able to generalize both to novel arrangements and novel object classes, achieving high performance across all four organizational schemas we tested.
\end{itemize}

We compare ConSOR with two baselines, a collaborative filtering-based approach to grouping objects based on learned pairwise similarity scores \cite{abdo_organizing_2016} and the GPT-3 large language model \cite{brown2020language}, on a withheld set of novel object arrangements and object types.  Our results show that ConSOR strongly outperforms both baselines in every tested category, without assuming that the target organization scheme is known \textit{a priori}.

\section{Related Work}

Numerous works in the literature have proposed approaches to goal-conditioned object rearrangement.  The means by which a user specifies the goal varies.  In some works, the goal is represented by a set of logical predicates encoding relationships between objects~\cite{paxton2022predicting, wang2022generalizable}. Alternately, in \textit{visual object rearrangement}, the goal is specified as an image, and the robot must perform object matching between the initial and goal images to determine the required object placement~\cite{goyal2022ifor, wu2022transporters, goodwin2022semantically}. 
A third form of goal specification is \textit{natural language instruction}, and recent work in language-conditioned manipulation has contributed techniques that ground a language description of the desired goal to the observable environment while performing zero-shot generalization to novel language commands~\cite{liu2022structdiffusion, shridhar2022perceiver, shridhar2022cliport, jiang2022vima}. 
Critically, all of the above methods are ineffective in the absence of an explicit goal specification.

To perform rearrangement without goal specification, some recent works take the approach of learning user-specific preferences, often modeling these preferences from a single demonstration~\cite{jain_transformers_2022, kapelyukh_my_2022}. These methods translate preferences encoded in the user demonstration, such as the order of moving objects or a preferred location of an object category, to a novel environment in a zero-shot manner, thereby eliminating the need to constantly provide task instructions. However, the above methods do not model object similarities, thereby limiting the scope of these approaches to template-like arrangement tasks (e.g., table setting, arranging an office desk). Additionally, these methods still require a demonstration for every new user or preference style.

Closest to our work is the collaborative filtering technique by Abdo et al. that learns user-specific preferences of grouping objects in containers as pairwise similarities between object categories~\cite{abdo_organizing_2016}. This technique is also extended to both rearrange novel object categories and optimize probing for the preferences of a new user. However, the agent in Abdo et al. assumes that the type of organization being sought is known \textit{a priori} (e.g., the robot knows to organize items by class, or by action affordances). Thus, for example, in putting away groceries or organizing a shelf, the user would always be required to specify the target organization type.  Our work relaxes this assumption and does not assume the organization type is known \textit{a priori}.  Instead, our model infers the desired object similarities from contextual cues in the observed initial state. Additionally, the approach by Abdo et al. is limited to approaching rearrangement as modeling pairwise similarities between object instances while our proposed framework can model more general semantic similarities between objects. 
Finally, their method requires a mixture of experts in order to generalize to novel object categories while the proposed ConSOR framework only needs the ConceptNet labels of novel objects to perform zero-shot generalization.

\begin{figure*}[ht]
    \centering
    \includegraphics[width=0.8\textwidth]{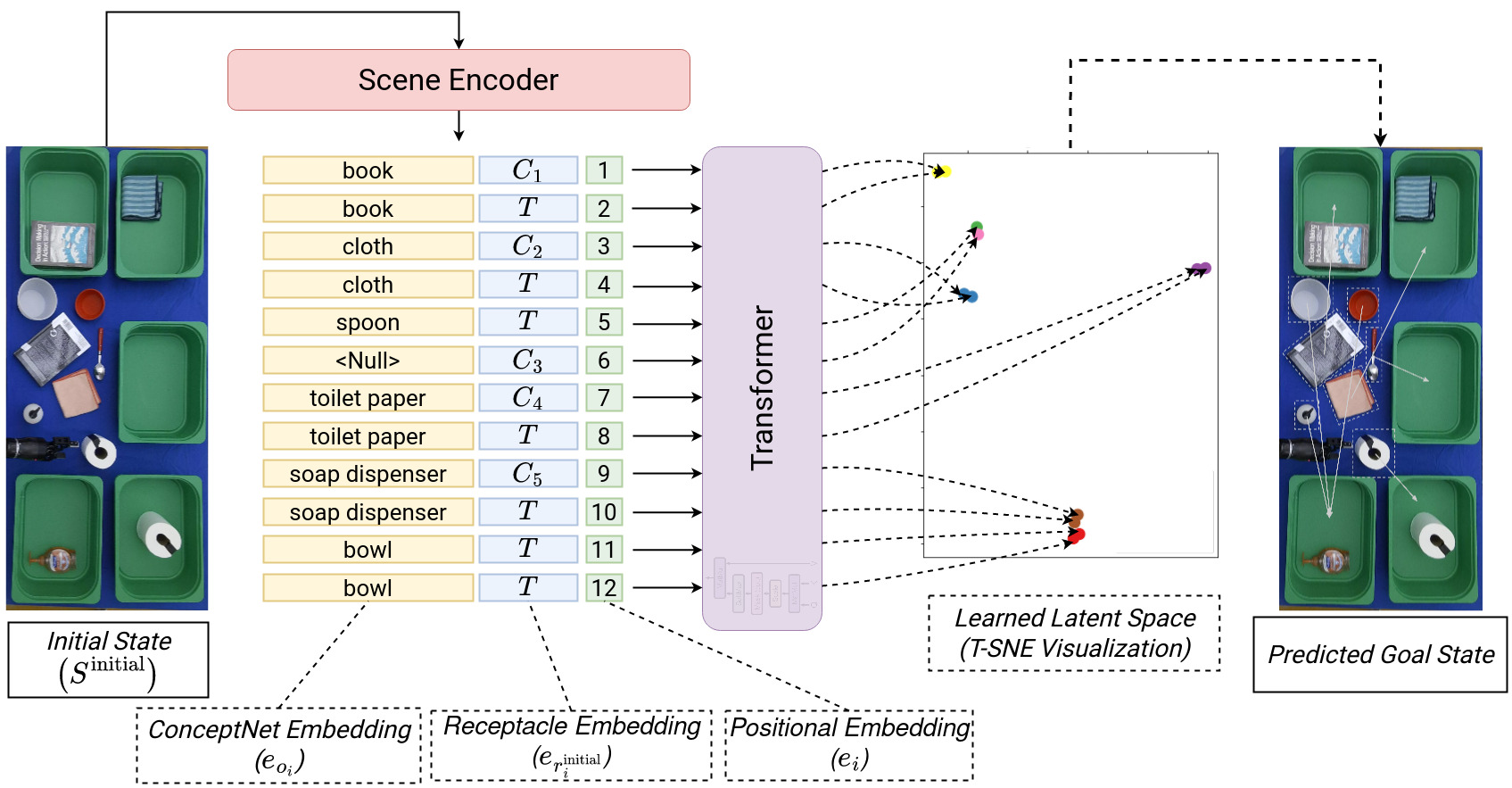}
    \caption{Model architecture showing how ConSOR encodes a rearrangement scene and transforms it into a learned embedding space, which is then used to determine object placements in the predicted goal state. Each green bin represents a container receptacle $c_i \in \mathcal{C}$. In the initial state, $\mathcal{X}_P$ is the set of objects already in a green bin, and $\mathcal{X}_U$ is the set of objects on the blue table. (right) Predicted receptacle assignments for all objects in $\mathcal{X}_U$ are shown using white arrows.}
    \label{fig:model-arch}
\end{figure*}

Another approach to object rearrangement is to learn user-agnostic object placement preferences from crowdsourced object arrangements.
Toris et al. present a multi-hypothesis model to learn to pick and place task templates representing the preferred placement locations of objects from human demonstrations ~\cite{toris_unsupervised_2015}. 
Sarch et al. propose an embodied AI rearrangement framework that learns from commonsense object-location preferences in tidy households to identify misplaced objects in a novel home environment and move the object to the best matching receptacle~\cite{sarch_tidee_2022}. 
In a similar work, Kant et al. contribute a benchmark and baseline for tidying household environments by identifying and rearranging misplaced objects without instruction using commonsense knowledge derived from a large language model~\cite{kant2022housekeep}. 
Though the above rearrangement approaches successfully avoid task goal specifications, the methods only focus on modeling object-receptacle preferences and do not reason about pairwise object similarities when placing objects. 
In a different work, Wu et al. propose an imitation learning framework to learn the target distribution of desired object arrangements from expert examples as a gradient field ~\cite{wu_targf_2022}. The learned target gradient field can then be used as a reward function to train a Reinforcement Learning (RL) agent to rearrange objects. However, the approach by Wu et al. requires separate models for different goal arrangement distributions and cannot identify the desired target distribution from the initial environment state, thereby restricting its usage to rearrangement tasks with only a single organizational style. 

\begin{figure*}[t]
    \centering
    \includegraphics[width=0.825\textwidth]{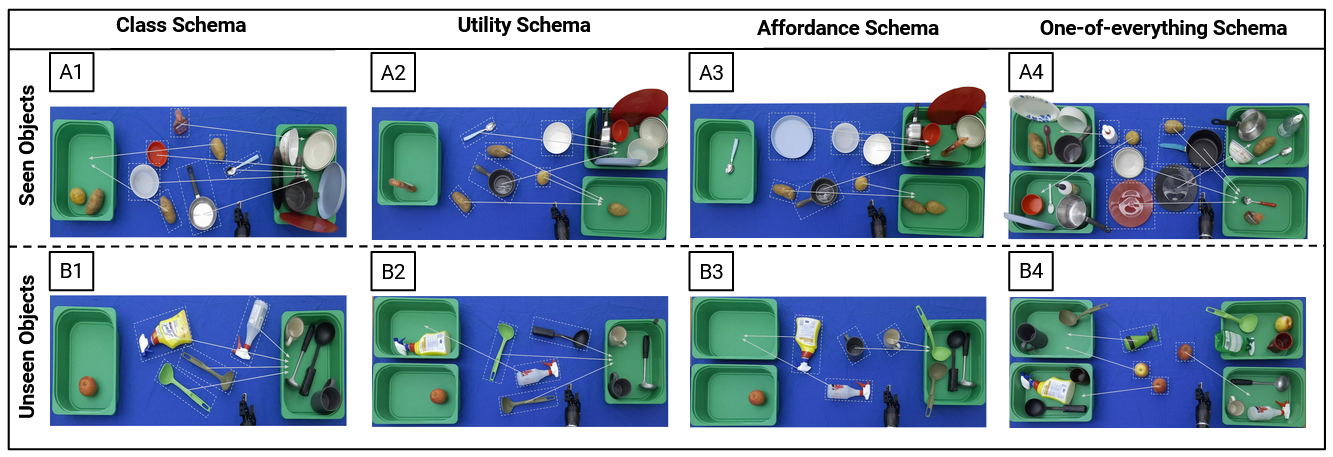}
    \caption{Goal states for each schema from two different sets of objects. The top row of scenes are from the test dataset with seen object categories, while the bottom row of scenes are from the test dataset with unseen object categories.}
    \label{fig:dataset}
\end{figure*}
\section{Object Rearrangement of Partially Arranged Environments}

We formalize object rearrangement in partially arranged environments as an instance of the general class of object rearrangement problems \cite{batra2020rearrangement} in which the goal arrangement state is not explicitly stated to the robot, but instead must be inferred from the context provided by already arranged items in the scene.  Specifically, the robot is presented with:
\begin{itemize}
    \item a fixed set of receptacles $\mathcal{R}$ in which objects can be placed (e.g., shelves, bins, containers),
    \item a set of prearranged objects, $\mathcal{X}_P$, arranged within the receptacles to match the user's intended organization schema (e.g., partially filled pantry where items are separated by meal type), and
    \item a set of unarranged objects, $\mathcal{X}_U$, for which the robot must find the correct receptacle in order to match the user's organizational schema.
\end{itemize}
Note that the user's schema is not explicitly specified and must be inferred from $\mathcal{X}_P$ and $\mathcal{R}$.  Multiple schemas can be applied to the same set of objects, and the robot's challenge is to infer the correct one and place $\mathcal{X}_U$ accordingly.

At any given time, we model the state $\mathcal{S}$ of the rearrangement environment as a set of tuples $\{x_i, \dots, x_{N_x}\}$, where $N_x$ is the number of object instances in $\mathcal{S}$. Each object instance is represented as $x_i = (o_i, r_i, i)$, where $o_i$ is the object category/class, $r_i \in \mathcal{R}$ is the receptacle in which $o_i$ is placed, and $i$ is an identifier index to distinguish object instances of the same category (e.g., multiple bowls in the same scene are assigned different values of $i$ in the state representation $\mathcal{S}$). 
$\mathcal{R}$ contains a work surface $T$ and the set of immovable containers $\mathcal{C} = \{C_i, \dots C_{N_C}\}$, where $N_C$ is the total number of containers in $\mathcal{S}$. Note that the containers in $\mathcal{C}$ can generalize to different cabinets, shelves, or drawers in a real household.
We represent a receptacle $r_j$ that does not contain any object in $\mathcal{S}$ by artificially placing a `null' object in the receptacle and adding it to the state.

Given a partially arranged initial state $\mathcal{S}^{initial}$, our goal is to reach a desired goal state $\mathcal{S}^{goal}$ by moving objects on $T$ (i.e., objects in $\mathcal{X}_U$) to containers $\mathcal{C}$ such that the resulting arrangement matches the user's latent organizational schema.  Note that our current problem does not consider visual semantics such as appearance similarities; however, this formulation can be extended to consider visual features by adding observation information to the state representation.

\section{ConSOR Transformer Model Architecture}

To address the above problem, we introduce the Context-aware Semantic Object Rearrangement framework (ConSOR). ConSOR uses a learned Transformer encoder to generate an object-centric latent embedding space from the partially arranged initial state that mimics the object grouping in the desired goal state. The object-centric embeddings are then clustered to determine object placements in the predicted goal state. 
Prior work in robot rearrangement has combined object-centric state representations with the attention capabilities of Transformer encoders to enhance the generalization capabilities of these models to novel objects, scenes and tasks~\cite{jain_transformers_2022, jiang2022vima, liu2022structformer}. We adopt a similar approach in designing the encoder model of ConSOR and augment object-centric representations with commonsense knowledge from ConceptNet to generalize to novel object categories.   

Figure~\ref{fig:model-arch} presents a detailed structure of the ConSOR framework. Given the initial state $\mathcal{S}^{initial}$, we project each object instance $x^{initial}_i=(o_i, r^{initial}_i, i)$ to a higher-dimensional space using a scene encoder $h(x^{initial}_i) \rightarrow e_i$, where $e_i = [e_{o_i}, e_{r^{initial}_i}, e_{i}]$. Specifically, $e_{o_i}$ is the pre-trained ConceptNet Numberbatch vector~\cite{speer2017conceptnet} corresponding to the category $o_i$, $e_{r^{initial}_i}$ is a positional encoding to indicate which receptacle the object lies in, and $e_{i}$ is a positional encoding of the indicator index $i$.
% Training and Evaluation
The encoded scene $\mathcal{V}^{initial}= \{h(x^{initial}_i), \dots h(x^{initial}_{N_x})\}$ is passed through ConSOR's Transformer encoder to output normalized latent embeddings $\mathcal{L} = \{l_i, \dots l_{N_x}\}$. The encoder is trained with a triplet margin loss~\cite{dong2018triplet} to group embeddings of objects sharing the same receptacle in the goal state together and move embeddings of objects in different receptacles away from each other.
In this manner, the encoder learns to generate a latent embedding space from the partially arranged initial state that mimics the object grouping in the desired goal state.
During evaluation, ConSOR chooses the container to place each unarranged object instance $x_i^{U} \in \mathcal{X}_{U}$ by calculating the centroid of each container in the latent space and choosing the container whose centroid has the highest cosine similarity with the corresponding latent vector $l_i^U$. Mathematically, the predicted placement $\hat{r}_i^U$ of instance $x_i^U$ is determined as
\begin{equation}
    \hat{r}_i^{U} = \argmax_{c \in \mathcal{C}} l_i^{U} \cdot l_c^{centroid}  
\end{equation}
\noindent where $l_c^{centroid}$ is the latent centroid embedding of container $c$ in $\mathcal{S}^{initial}$. 

The encoder model of ConSOR consists of three stacked Transformer encoder layers, followed by an MLP layer to reduce the dimension of the generated embeddings and a $L_2$ normalization layer. We train the model for $30$ epochs (learning rate=$1e-3$, batch size=$64$, dropout=$0.5$) and perform early stopping based on the success rate obtained from evaluating on the validation dataset. 

\section{Dataset of Organizational Schemas for Object Rearrangement}
To evaluate object rearrangement in partially arranged environments, we contribute a novel dataset of arranged scenes generated using household objects from the AI2Thor simulator~\cite{Kolve2017AI2THORAI}. To generate each scene, we defined four organizational schemas to determine how objects are grouped in the goal state:
\begin{enumerate}
    \item Class schema ($F_{class}$): grouping objects based on the affinity of their semantic concepts in WordNet~\cite{miller1995wordnet},
    \item Utility schema ($F_{utility}$): grouping objects based on the affinity of their product categories mined from a popular retail store (Walmart~\cite{walmart}),
    \item Affordance schema ($F_{affordance}$): grouping objects with similar action affordance labels (6 affordance labels in total) gathered from the Moving Objects dataset~\cite{goyal2017something}, and
    \item One-of-everything ($F_{OOE}$): distributing objects in containers such that each container holds exactly one of each object type (referred to as OOE for brevity). 
\end{enumerate}

We created a schema-balanced dataset by generating $1980$ training, $110$ validation, and $110$ test goal scenes from each of the four schemas using a set of $28$ object categories taken from the AI2Thor simulator~\cite{Kolve2017AI2THORAI} and grounded in WordNet. 
Example object categories include fruits, vegetables, office supplies, kitchen and dining accessories, cleaning supplies, bathroom accessories, and home decor.
We also generated a secondary test dataset of $120$ goal scenes using $10$ novel object categories to test the generalization capability of ConSOR to object categories that were entirely unseen during training. Table~\ref{tab:DatasetObjects} lists the objects present in our dataset. 
Partially arranged initial scenes are generated from goal scenes by sequentially removing randomly selected objects in $\mathcal{X^P}$ from containers and placing them on work surface $T$. In this manner, we systematically vary the degree to which the presented organization is complete.  

Figure~\ref{fig:dataset} shows example arranged scenes from each schema. In the Class Schema (A1 and B1), objects are grouped by class similarity, such that vegetables should be placed in one bin, and kitchen items in the other.  Note that in B1, the model is asked to generalize to cleaning supplies, with the goal of grouping them with kitchen items rather than vegetables due to more closely aligned similarity, as only two containers are provided in this example.

In the Utility Schema (A2, B2), the robot is provided with three containers. In A2, vegetables, a soap dispenser, and cooking supplies are organized into different bins.  In B2, the model generalizes to previously unseen objects, placing cleaning supplies in their own bin.  

In the Affordance Schema (A3, B3), objects are grouped by their afforded functionality. As this example highlights, detecting the desired organizational structure from a partial scene can be quite challenging.  The key clue is given by the spoon, which is placed separately from other kitchen items. This is due to the spoon's shape (long handle with shallow convex hull) differing from that of the other objects (round with deep convex hull), thereby resulting in different affordances.  Thus, the robot must learn to appropriately group the remaining items. Furthermore, note that scenes B2 and B3 have different initial states but end up in the same goal state; this type of aliasing makes the partial rearrangement problem complex, causes the differences between schemas to be less obvious, and requires that the learned model pay close attention to contextual cues in the initial state.

Finally, in the One-of-Everything Schema (A4, B4), the robot's objective is to place one of each object in the bins, akin to packing a lunch, or conference gift bags.  We include this schema because, while appearing simple, it can pose quite a challenge to machine learning algorithms because similar objects must be separated rather than binned together.  As we will show in the results, prior works struggle to find solutions to this schema.

Note that, although we define four types of schemas, ConSOR is trained only on the initial and goal states without any schema labels. Instead, ConSOR learns to distinguish between scenes of different schemas by learning the differences between them from the training data.

\begin{table}[]
    \centering
    \begin{tabular}{ | p{\dimexpr 0.6\linewidth-2\tabcolsep} |
                       p{\dimexpr 0.4\linewidth-2\tabcolsep} | } \hline
    Seen Object Categories & Unseen Object Categories \\ \hline
    Aluminum Foil, Basket Ball, Book, Bottle, Bowl, Bread, Candle, Cloth, Cup, Dish Sponge, Dumbbell, Egg, Hand Towel, Kettle, Laptop, Lettuce, Newspaper, Pen, Plate, Pot, Potato, Scrub Brush, Soap Dispenser, Spoon, Toilet Paper, Tomato, Towel, Wine Bottle &  Apple, Box, Ladle, Mug, Pan, Paper Towel Roll, Pencil, Spray Bottle, Vase, Watering Can \\ \hline
    \end{tabular}     
    \caption{Table of $28$ object categories seen (left) and $10$ unseen (right) during training}
    \label{tab:DatasetObjects}
    \vspace{-2.5em}
\end{table}

\begin{table*}[t]
    \centering
    \begin{tabular}{|c|c|c|c|c|c|c|c|c|}
    \hline
    Method & \multicolumn{2}{|c|}{Class Schema ($F_{class}$)} & \multicolumn{2}{c|}{Utility Schema ($F_{utility}$)} & \multicolumn{2}{|c|}{One-of-everything Schema ($F_{OOE}$)} & \multicolumn{2}{c|}{Affordance Schema ($F_{affordance}$)} \\
    \hline
    & $M^{SR}$ & $M^{NSED}$ & $M^{SR}$ & $M^{NSED}$ & $M^{SR}$ & $M^{NSED}$ & $M^{SR}$ & $M^{NSED}$ \\
    \hline
    ConSOR (\textbf{Ours}) & 
    \textbf{99\%} & \textbf{1.4 (SD=0.5)} & \textbf{99\%} & \textbf{1.2 (SD=0.4)} &
    \textbf{100\%} & - & \textbf{98\%} & \textbf{1.0 (SD=0.0)} \\
    \hline
    \textit{Abdo-CF}& 89\% & 3.9 (SD=1.3) & 93\% & 3.6 (SD=0.8) & 0\% & 15.4 (SD=3.1) & 90\% & 3.4 (SD=1.1) \\
    \hline
    \textit{GPT-3} & 36\% & 3.1 (SD=2.1) & 41\% & 3.2 (SD=2.2) & 4\% & 9.8 (SD=5.6) & 40\% & 3.3 (SD=2.2) \\
    \hline % use \hhline{~|--|--|} for double line between rows
    \end{tabular}
    \caption{Evaluation results for each schema calculated on test data of rearrangement scenes with unseen object arrangements.}
    \label{tab:performance-unseen-obj-dist}
\end{table*}

\begin{table*}[t]
    \centering
    \begin{tabular}{|c|c|c|c|c|c|c|c|c|}
    \hline
    Method & \multicolumn{2}{|c|}{Class Schema ($F_{class}$)} & \multicolumn{2}{c|}{Utility Schema ($F_{utility}$)} & \multicolumn{2}{|c|}{One-of-everything Schema ($F_{OOE}$)} & \multicolumn{2}{c|}{Affordance Schema ($F_{affordance}$)} \\
    \hline
    & $M^{SR}$ & $M^{NSED}$ & $M^{SR}$ & $M^{NSED}$ & $M^{SR}$ & $M^{NSED}$ & $M^{SR}$ & $M^{NSED}$ \\
    \hline
    ConSOR (\textbf{Ours}) & 
    \textbf{99\%} & \textbf{1.0 (SD=0.0)} & \textbf{91\%} & \textbf{1.3 (SD=0.5)} & 
    \textbf{100\%} & - & \textbf{97\%} & \textbf{1.0 (SD=0.0)} \\
    \hline
    \textit{GPT-3} & 
    40\% & 3.3 (SD=2.2) & 61\% & 2.9 (SD=2.2) & 
    3\% & 9.2 (SD=5.4) & 44\% & 3.2 (SD=2.1) \\
    \hline 
    \end{tabular}
    \caption{Evaluation results for each schema calculated on test data of rearrangement scenes with novel object categories.}
    \vspace{-1em}
    \label{tab:performance-unseen-objs}
\end{table*}

\section{Baselines and Metrics}
We compare ConSOR performance against two baselines:

\smallskip
\noindent \textit{\textbf{Abdo-CF}}: \textit{Abdo-CF} is a collaborative filtering technique proposed by Abdo et al. that learns user-specific pairwise object similarities from multiple preference ranking matrices of different users~\cite{abdo_organizing_2016}. 
The learned pairwise preferences per user are then used to identify the placements of query objects via spectral clustering. Critically, this approach requires that the organization or schema type be known \textit{a priori} and given as an input.  By comparison, ConSOR implicitly infers the schema from contextual cues in the partially arranged initial scene. Additionally, \textit{Abdo-CF} generalizes to unseen objects by relying on a mixture of experts providing object similarities of the unseen object, while ConSOR only requires the ConceptNet labels of unseen objects. 

\begin{figure}
    \centering
    \includegraphics[width=0.44\textwidth]{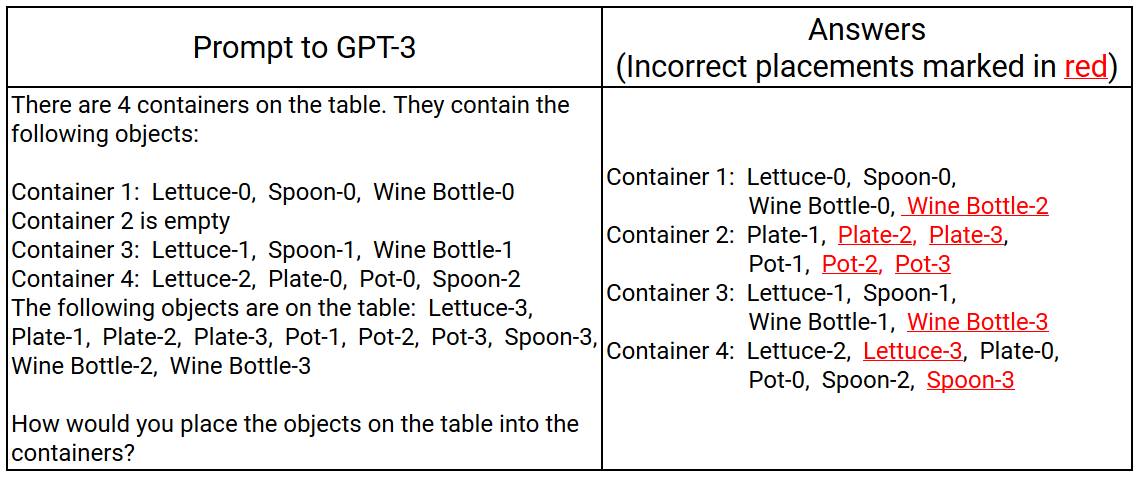}
    \caption{Prompt given to GPT-3 and its response for a one-of-everything schema scene. The misplaced objects are marked in red underline.}
    \vspace{-1.3em}
    \label{fig:gpt-3example}
\end{figure}

\smallskip
\noindent \textbf{\textit{GPT-3}}: The Generative Pre-trained Transformer 3 (\textit{GPT-3}) is an autoregressive language model that generates natural language in response to user input~\cite{brown2020language}. Prior works have demonstrated that large language models such as \textit{GPT-3} are able to reason about sequential tasks and physical spaces~\cite{logeswaran-etal-2022-shot, kirk2022evaluating, brohan2022can}.  To utilize \textit{GPT-3} as a baseline, we prompt the model with a description of the partially arranged initial state along with one unlabeled demonstration from each schema to inform the language model about the desired output. 
Figure~\ref{fig:gpt-3example} shows a prompt taken from the test dataset and the corresponding response from GPT-3. 

%\subsection{Evaluation Metrics}
In our problem formulation, we seek to transform an initial object arrangement, represented by state $\mathcal{S}^{initial}$, to a goal object arrangement, represented by the goal state $\mathcal{S}^{goal}$, where the goal is not know to the robot \textit{a priori} and must be inferred.  We therefore evaluate object arrangement performance by measuring the similarity between the achieved object arrangement state and the goal state.  To quantify this difference, we introduce a distance measure derived from \textit{edit distance}, a widely used string similarity metric in computational linguistics~\cite{levenshtein1966binary}. Specifically, we define the Scene Edit Distance ($SED$) between states $\mathcal{S}^A$ and $\mathcal{S}^B$ as the minimum number of object displacements that must be made in $\mathcal{S}^A$ to reach $\mathcal{S}^B$. In our problem formulation, this is equivalent to the number of misplaced objects in $\mathcal{S}^A$ compared to $\mathcal{S}^B$ and vice-versa.

Additionally, we derive two aggregate evaluation metrics from $SED$ to measure rearrangement performance across an entire dataset.  The first is the Success Rate, $M^{SR}$, which corresponds to the fraction of goal states predicted correctly:
\begin{equation}
    M^{SR} = \frac{1}{D}\cdot \sum_{i=1}^{D} \mathds{1}(SED(\hat{\mathcal{S}}_i, \mathcal{S}^{goal}_i) = 0)
\end{equation}
where $\hat{\mathcal{S}}_i$ is the predicted goal state, $\mathcal{S}^{goal}_i$ the ground truth state for the initial state $S^{initial}_i$, and $D$ is the total number of examples in the test dataset.

The second metric is the Average Non-zero SED $M^{NSED}$ or the average $SED$ between incorrectly predicted goal states ($SED > 0$) and their ground truth states. This is defined as:

\begin{equation}
    M^{NSED} = 
    \frac{\sum_{i=1}^{D} SED(\hat{\mathcal{S}}_i, \mathcal{S}^{goal}_i)\cdot 
            \mathds{1}(SED(\hat{\mathcal{S}}_i, \mathcal{S}^{goal}_i) > 0)}
        {\sum_{i=1}^{D} \mathds{1}(SED(\hat{\mathcal{S}}_i, \mathcal{S}^{goal}_i) > 0)}
\end{equation}
Together, the above two metrics capture a model's performance, such that $M^{SR}$ reports the percentage of arrangements that an algorithm gets completely right, and $M^{NSED}$ reports the degree of dissimilarity for scenes that were not correct (non-zero SED).

\section{Evaluation Results}
In this section, we present results of two generalization experiments, first evaluating generalization to previously unseen arrangements with known objects, and second evaluating zero-shot generalization to novel object categories.  Additionally, we present insights characterizing the differences in learned embedding spaces across schemas, and evaluate the effect of training data size on performance.

\begin{figure}[t]
    \centering
    \includegraphics[width=0.35\textwidth]{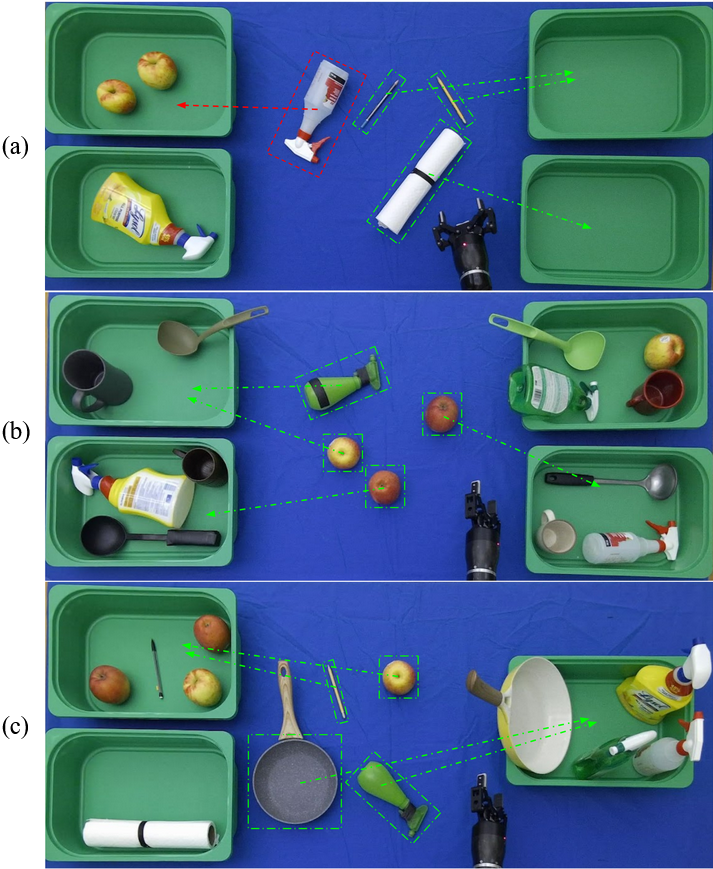}
    \caption{Predicted goal states generated by ConSOR for scenes with novel object categories. Correct object placements are shown using a dash-dotted green bounding box and arrow, while incorrect object placements are shown using a dashed red bounding box and arrow.}
    \label{fig:visualization-unseen-objs}
    \vspace{-1em}
\end{figure}

\subsection{Generalizing to Unseen Object Arrangements}

Table~\ref{tab:performance-unseen-obj-dist} presents a summary of evaluation results from testing on scenes with unseen object arrangements. ConSOR yields a higher $M^{SR}$ than both \textit{Abdo-CF} and \textit{GPT-3} across all four schemas. Notably, our framework is also the only method in our evaluation to perfectly rearrange $F_{OOE}$ scenes. Additionally, ConSOR has the least $M^{NSED}$ score across all four schemas, indicating that, in the rare cases that errors occur, ConSOR generates state predictions that are closer to the true goal state than the baseline approaches. 
\textit{Abdo-CF} has the second-best performance in three out of four of the schemas while failing to successfully rearrange a single $F_{OOE}$ scene. We attribute the failure in $F_{OOE}$ by \textit{Abdo-CF} to the inductive bias of collaborative filtering, as it is difficult to mimic the $F_{OOE}$ schema using pairwise object similarities. \textit{GPT-3} performs the worst on three out of four schemas with a slightly higher $M^{SR}$ in $F_{OOE}$ than \textit{Abdo-CF}. This shows that the general-purpose commonsense knowledge learned by \textit{GPT-3} is insufficient to 
model an organizational schema with a specific set of semantic constraints, thus necessitating the need for our proposed framework.

\begin{figure*}[t]
    \centering
    \includegraphics[width=0.78\textwidth]{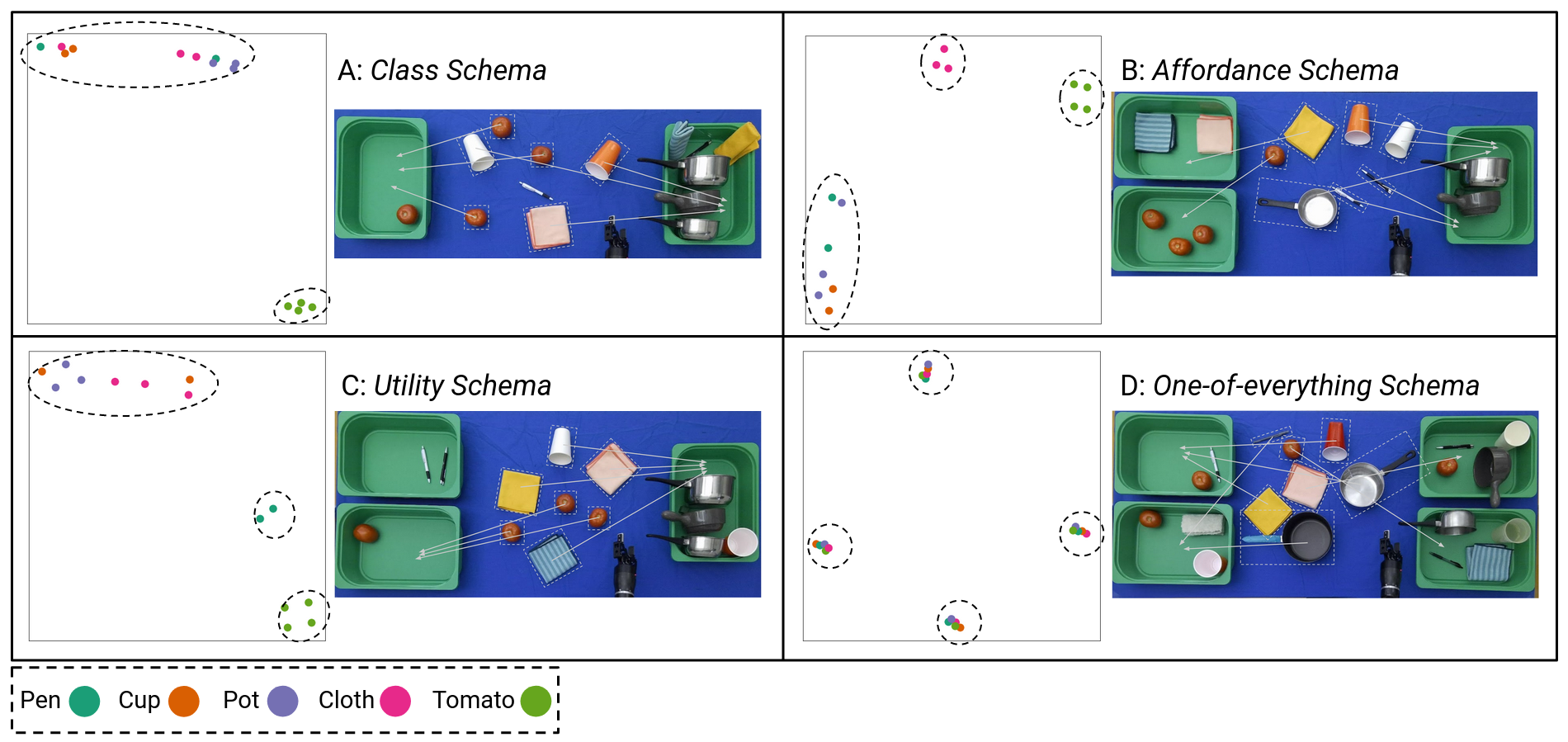}
    
    \caption{Visualizing the learned embedding space of ConSOR for scenes of different schemas. The right side of each cell shows the initial scene and ground truth object placements, and the left side is a T-SNE projection of the generated object-centric embeddings in two dimensions.}
    \label{fig:plot-tsne-viz}
\end{figure*}

\begin{figure*}[t]
    \centering
    \includegraphics[width=0.78\textwidth]{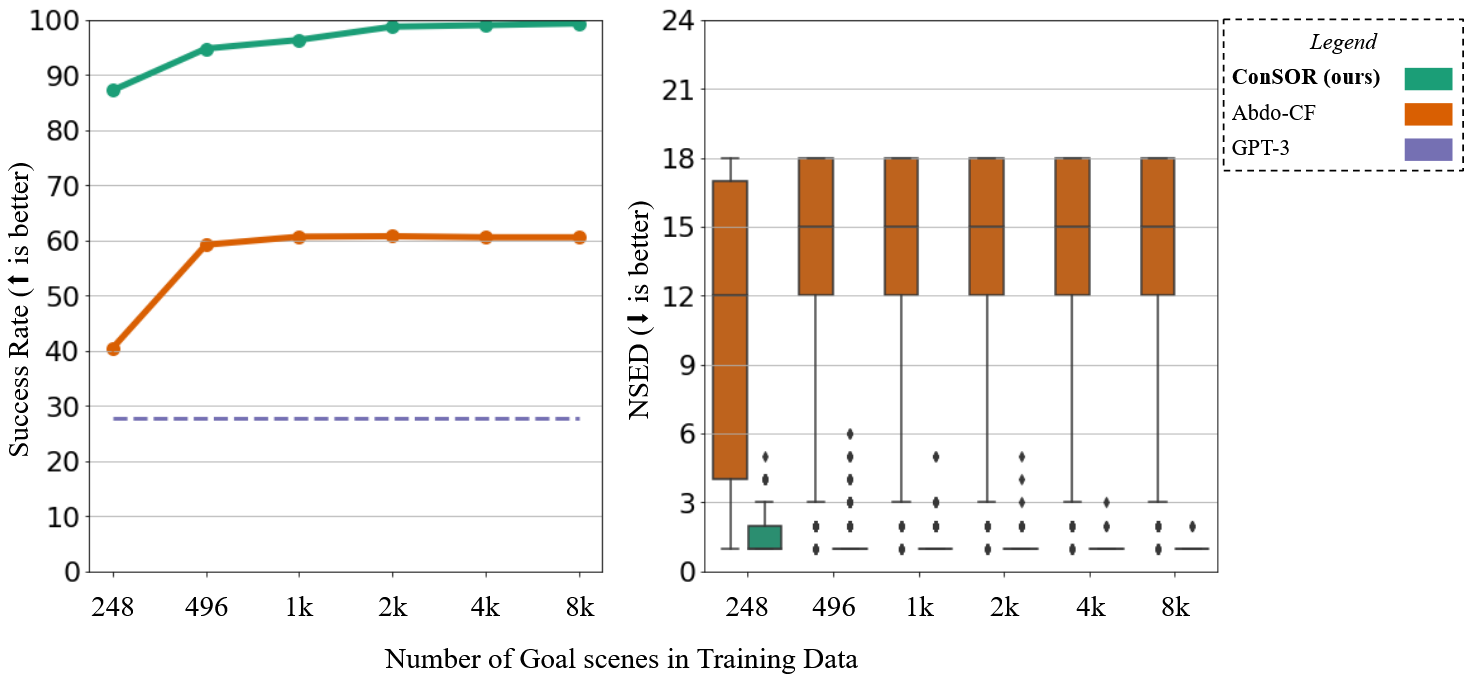}
    \caption{Measuring the change in performance of each rearrangement method based on the number of goal scenes in the training data. The plot on the left is a line diagram of success rate averaged across all schemas versus the size of training data, with the success rate of \textit{GPT-3} shown as a dotted line. The plot on the right is a series of box plots showing the varying range of non-zero $SED$ scores for ConSOR and \textit{Abdo-CF} with an increase in training data.
    }
    \label{fig:plot-perf-vs-train-data}
\end{figure*}

\subsection{Zero-Shot Generalization to Novel Object Categories}

Table~\ref{tab:performance-unseen-objs} shows our evaluation results from testing on scenes with novel object categories. We do not evaluate the baseline \textit{Abdo-CF} on novel object categories as this method requires external semantic knowledge to rearrange objects unseen during training, and the lack of this knowledge with other methods leads to an unfair comparison.
ConSOR is able to successfully leverage the commonsense knowledge embedded in ConceptNet to perform zero-shot generalization to completely novel scenes and outperform \textit{GPT-3}. Also, in comparison to our model's performance on unseen object arrangements, ConSOR retains performance on three out of four schemas, with $F_{utility}$ showing the largest drop in performance. We believe this drop in performance is due to the Utility schema deviating the most from the commonsense knowledge embedded in ConceptNet.

Figure~\ref{fig:visualization-unseen-objs} presents some of the correct and incorrect goal predictions made by ConSOR for scenes with novel object categories. We observe in Figure~\ref{fig:visualization-unseen-objs}(a) that ConSOR occasionally places objects of the same category in separate containers even when the desired goal schema is not $F_{OOE}$. We hypothesize that this may be attributed to the model lacking confidence about the desired schema, resulting in a goal state with a `hybrid' schema. Figures \ref{fig:visualization-unseen-objs}(b) and \ref{fig:visualization-unseen-objs}(c) show two accurate goal predictions, belonging to $F_{OOE}$ and $F_{utility}$ respectively. In both, ConSOR leverages contextual cues about the desired schema from the initial state, such as the number of containers in the scene and the current object arrangement, to perfectly generate the goal state.

\subsection{Visualizing the difference in scene organization across schemas in learned embedding space}
Evaluation results in previous sections show that ConSOR successfully learns a mapping from the initial partially arranged rearrangement state to a latent space of embeddings mimicking the desired object grouping. We visualize the embeddings generated by our method for four different initial state configurations from the test dataset in Figure~\ref{fig:plot-tsne-viz} using a T-SNE projection~\cite{van2008visualizing}. We observe that, for the same set of object categories, the learned embedding space of ConSOR adapts itself to the structure of the scene (number of containers) as well as the organization of objects in the initial state. For example, the latent spaces of scenes A (Class schema) and D (One-of-everything Schema) highlight the differences in the two scenes, namely the different numbers of containers in the scene and whether the same object categories are grouped together or separately. On the other hand, scenes B (Affordance schema) and C (Utility schema) can only be differentiated by observing whether `Pen' or `Cloth' is grouped with `Pot', and this is seen in the latent space as well. 
We also note that ConSOR is able to identify the differences between scenes of different schemas without being explicitly trained with the actual schema labels, and instead learns to differentiate schemas from unlabelled training data.

\subsection{Effect of Size of Training Data on Performance}

Finally, we evaluate the effect of training data size on the performance of ConSOR and \textit{Abdo-CF}. Figure~\ref{fig:plot-perf-vs-train-data} shows the change in average $M^{SR}$ and $M^{NSED}$ values for different numbers of goal scenes in the training data. All metrics were calculated using the same test dataset as the previous sections. We observe the average $M^{SR}$ of ConSOR steadily rising and the variance of $M^{NSED}$ scores decreasing with more training data, while \textit{Abdo-CF} performance saturates after $496$ training goal scenes. Across all training data sizes, we find that ConSOR has a higher average $M^{SR}$ and lower mean $M^{NSED}$ score than \textit{Abdo-CF}.  

\section{Conclusion and Discussion}

This work introduces ConSOR, a semantic reasoning framework for object rearrangement. ConSOR relies on contextual cues from a partially arranged environment to infer the desired goal state by generating a learned object-centric latent space that mimics the arrangement in the desired goal state. Additionally, ConSOR leverages external commonsense knowledge from ConceptNet to perform zero-shot generalization to rearrange scenes with novel object categories. We evaluated our proposed framework on a dataset of~8k arranged scenes, each belonging to one of four `organizational schemas', and found that our approach strongly outperforms both the \textit{Abdo-CF} and \textit{GPT-3} baseline across all tested conditions.  

Note that ConSOR outperforms the next leading baseline, \textit{Abdo-CF}, even though the baseline is explicitly given the target schema type as input (e.g., \textit{class schema}) while ConSOR is required to infer this information automatically from context.  In a real-world setting, ConSOR would therefore require significantly less effort from the user.

One assumption of our approach is that the robot knows which items have been pre-arranged, and thus serve as useful context, and which still need to be put away. Depending on the real-world application, making this distinction may be simple or quite difficult (e.g., bags of new groceries are easy to detect, out-of-place living room objects pose a greater challenge). Recent work on visuo-semantic commonsense priors may help inform this decision in the future \cite{sarch_tidee_2022, kant2022housekeep}.

\bibliographystyle{IEEEtran}
\bibliography{bibliography.bib}

\end{document}